\documentclass[a4paper, 12pt]{article}

\usepackage[a4paper, left=3cm, top=3cm, right=3cm, bottom=3cm ]{geometry}
\usepackage{authblk}
\usepackage[onehalfspacing]{setspace}

\usepackage[T1]{fontenc}

\usepackage{cite}
\usepackage{amsmath,amssymb,amsfonts}
\usepackage{algorithmic}
\usepackage{graphicx}
\usepackage{algorithm,algorithmic}
\usepackage{hyperref}
\hypersetup{hidelinks=true}
\usepackage{textcomp}

\usepackage{breakurl}

\usepackage{siunitx}

\usepackage{array}
\usepackage{stfloats}
\usepackage{url}
\usepackage{verbatim}
\usepackage{siunitx}
\usepackage{graphicx}

\usepackage{multirow}
\usepackage{subcaption}

\usepackage{todonotes}

\usepackage{siunitx}

\usepackage{mathtools}
\usepackage{booktabs}
\usepackage{dsfont}
\usepackage{pgfplots}
\usetikzlibrary{intersections, pgfplots.fillbetween}
\definecolor{mycolor}{RGB}{85,125,250}
\definecolor{gcolor}{RGB}{0.01,0.199,0.1}

\newcommand{\abs}[1]{\lvert#1\rvert}
\newcommand{\norm}[1]{\lVert#1\rVert}

\newcommand{\fixed}{f}
\newcommand{\moving}{m}
\newcommand{\warped}{w}
\newcommand{\disp}{\boldsymbol{u}}
\newcommand{\deform}{\boldsymbol{\Phi}}
\newcommand{\ew}{\mathbb{E}}
\newcommand{\net}{\mathcal{U}}

\newcommand{\fun}{\mathcal{D}}
\newcommand{\gun}{\mathcal{R}}
\newcommand{\run}{\mathcal{L}}

\newcommand{\R}{\mathbb{R}}

\DeclareMathOperator{\lncc}{\mathcal{D}_{\mathrm{LNCC}}}
\DeclareMathOperator{\id}{Id}
\DeclareMathOperator*{\argmin}{arg\,min}

\numberwithin{equation}{section}
\numberwithin{figure}{section}

\newtheorem{theorem}{Theorem}[section]

\newtheorem{remark}[theorem]{Remark}

\numberwithin{equation}{section}
\numberwithin{figure}{section}
\numberwithin{theorem}{section}

\allowdisplaybreaks

\title{A lightweight residual network for unsupervised deformable image registration}

\date{June 14, 2024}

\author{Ahsan Raza Siyal}

\affil{Department of Mathematics, University of Innsbruck \authorcr
E-mail:  \texttt{ahsan.siyal@uibk.ac.at}
 }

\author{Astrid Ellen Grams}

\affil{Department of Radiology, Medical University of Innsbruck\authorcr
E-mail:  \texttt{astrid.grams@i-med.ac.at}
 }

\author{Markus Haltmeier}

\affil{Department of Mathematics, University of Innsbruck\authorcr
E-mail:  \texttt{markus.haltmeier@uibk.ac.at}
 }

\begin{document}

\maketitle

\begin{abstract} 

Accurate volumetric image registration is highly relevant for clinical routines and computer-aided medical diagnosis. Recently, researchers have begun to use transformers in learning-based methods for medical image registration, and have achieved remarkable success. Due to the strong global modeling capability, Transformers are considered a better option than convolutional neural networks (CNNs) for registration. However, they use bulky models with huge parameter sets, which require high computation edge devices for deployment as portable devices or in hospitals. Transformers also need a large amount of training data to produce significant results, and it is often challenging to collect suitable annotated data. Although existing CNN-based image registration can offer rich local information, their global modeling capability is poor for handling long-distance information interaction and limits registration performance. In this work, we propose a CNN-based registration method with an enhanced receptive field, a low number of parameters, and significant results on a limited training dataset. For this, we propose a residual U-Net with embedded parallel dilated-convolutional blocks to enhance the receptive field. The proposed method is evaluated on inter-patient and atlas-based datasets. We show that the performance of the proposed method is comparable and slightly better than transformer-based methods by using only $\SI{1.5}{\percent}$ of its number of parameters.


\end{abstract}

Deformable image registration plays a significant role in computer-aided diagnosis, medical image analysis, and daily clinical routine. The registration problem is typically formulated as a minimization problem that includes a similarity term, which measures the distance between a pair of fixed and warped moving images, and a regularization term that penalizes non-regular deformations. Traditional methods use variational methods for finding a non-linear correspondence between the points of the two images~\cite{Avants2008,Vercauteren2009,Modat2010,Heinrich2013,scherzer2006mathematical}. However, due to the necessity to solve a complex optimization problem for each new pair of medical images, such methods are computationally expensive and typically slow in practice. A variety of iterative optimization techniques have been developed \cite{Avants2008,Vercauteren2009,Rueckert1999}, many of which have demonstrated excellent registration accuracy. However, such techniques are constrained by manual image pair adjustment and slow inference speeds. Recently, neural network-based registration has emerged as a strong standard for large-scale medical image registration \cite{Balakrishnan2018,Balakrishnan2019,Jia2022,tony,Qiu2022} because of their high inference rate and accuracy, which is equivalent and even better than iterative methods.

Deep neural networks (DNN), particularly convolutional neural networks (CNNs), have emerged as state-of-the-art in various computer vision applications, such as image segmentation~\cite{Long2015}, object recognition~\cite{Redmon2016}, and image classification~\cite{He2016} in recent years. Medical image analysis domains such as tumor segmentation~\cite{Isensee2020}, image reconstruction~\cite{Zhu2018}, and disease diagnostics have recently received a lot of attention with CNN-based approaches. In the context of deformable image registration, CNN-based approaches solve a single optimization problem during the training phase instead of per-image optimization for each registration approach. After training, the CNNs can quickly align an unknown image pair because they learn the standard representation of image registration from training images. Thus, they are much faster than variational and iterative methods while even producing noticeably better results.
Initial approaches on image registration with CNNs use supervision of ground-truth deformation fields, which are often produced using conventional registration techniques~\cite{Onofrey2014,Yang2017}. Due to the challenge of having a ground truth deformation field, recently the emphasis has switched to unsupervised techniques independent of ground-truth deformation fields~\cite{Balakrishnan2018,Balakrishnan2019,Dalca2019,Kim2021,Lei2020}. U-Net~\cite{Ronneberger2015} or variants of U-Nets are often  employed  in  existing deep-learning-based registration. Recently, transformer architecture~\cite{Liu2021,Chen2021,Chen2022} turned out to yield state of the art; however, these networks come at the cost of a large number of network parameters, the need for a large amount of training data, and numerically heavy training time.

In this study, we propose an efficient unsupervised 3D deformable network architecture for the registration of brain magnetic resonance images (MRIs). By comparing the global and local similarities of image pairings, we train our network in an unsupervised way without relying on ground-truth deformation fields or relevant anatomical labels. We suggest the residual U-Net with embedded parallel dilated-convolutional blocks to improve the receptive field. Atlas-based datasets and inter-patient datasets are used to evaluate the suggested technique. We demonstrate that the performance of the proposed method is similar or even better than transformer-based methods in limited training data scenarios by using only \SI{1.5}{\percent} of its parameters.

\textbf{Outline:} The remainder of this paper is organized as follows. Section~\ref{sec:intro} provides background on image registration and discusses related work. Section~\ref{sec:methods} explains the proposed methodology. Section~\ref{sec:results} discusses the implementation details, datasets used in this study, and experimental results. Finally, in Section~\ref{sec:conclusion}, we give a short summary of the paper and an outlook to future research.

\section{Image registration background}
\label{sec:intro}

The task of image registration is to transform one image into another such that they are well-aligned according to problem-specific criteria. Let us denote by  $\fixed \colon \Omega \to \R$ the fixed image and by $\moving \colon \Omega \to \R$ the moving image, both defined on the same domain $\Omega$. Mathematically, image registration involves finding a deformation $\deform \colon \Omega \to \Omega$ such that $\fixed = \moving \circ \deform$. In its basic form, this problem is clearly ill-posed, as it suffers from non-uniqueness and instability when $\fixed$ and $\moving$ contain noise. Non-uniqueness implies that there may be multiple deformations that can transform the moving image into the fixed image. Instability results in the deformation field being irregular and sensitive to changes in the provided image data. Therefore, the problem needs to be regularized \cite{scherzer2009variational}, meaning it should be approximated by a stable problem that incorporates prior information to account for noise in the data and non-uniqueness of the solution.

\subsection{Variational registration}

Traditional deformable image registration methods iteratively minimize an energy functional \cite{Avants2008,Heinrich2013,Ashburner2007,droske2004variational,sotiras2013deformable,scherzer2006mathematical}, resulting in an instance of a variational regularization method. In that context, the desired deformation $\deform$ transforming the moving image $\moving$ to the fixed image $\fixed$ is defined as a minimizer of the energy functional 
\begin{equation} \label{eq:tik}
	\run_{\fixed, \moving,\alpha} (\deform )  \triangleq \fun (\fixed, \moving \circ \deform )  +  \alpha\gun(\deform) \,.
\end{equation}  
Here $\fun$ is a similarity measure for quantifying the difference between the fixed image $\fixed$ and the deformed moving image $\moving \circ \deform$, $\gun$ is the regularization term penalizing infeasible deformations, and $\alpha$ is the regularization parameter that acts as a trade-off between matching the transformed moving image and the fixed image.

Various functionals $\fun$ and $\gun$ have been proposed in the literature. Some common choices for $\fun$ include normalized cross-correlation (NCC), mean squared error (MSE), structural similarity index (SSIM), or mutual information. The regularization term $\gun$ guarantees a regular deformation field. Some popular options are isotropic diffusion, anisotropic diffusion, total variation, and bending energy.  Often, the deformation is written in the form $\deform = \id + \disp$, where $\disp$ is the displacement vector field, and \eqref{eq:tik} is minimized within a space of displacement vector fields. These methods include statistical parametric mapping~\cite{Ashburner2000}, elastic-type methods~\cite{Bajcsy1989}, free-form deformations with B-splines~\cite{Rueckert1999}, and Demons~\cite{Vercauteren2009}.  Diffeomorphic transforms, which preserve topology, have achieved outstanding results in numerous computational anatomy research. Some of the famous formulations are Large Diffeomorphic Distance Metric Mapping (LDDMM)~\cite{Beg2005}, DARTEL~\cite{Ashburner2007}, and standard symmetric normalization (SyN)~\cite{Avants2008}.

\subsection{Unet-based registration}

More recent registration approaches  leverage ideas from unsupervised machine learning to determine the deformation field. These methods employ a network that takes stacked fixed and moving images as input and returns the displacement field as output. VoxelMorph~\cite{Balakrishnan2018, Balakrishnan2019} has been one of the first efforts in that direction. It uses a five-layer U-Net \cite{Ronneberger2015}  followed by three convolutional layers at the end to register medical images. The training of the network has been performed in an unsupervised manner, and for this, an unsupervised loss function was utilized, which consists of normalized cross-correlation as a similarity term and isotropic diffusion as a regularization term. VoxelMorph operates orders of magnitude faster and offers comparable accuracy to cutting-edge conventional approaches such as ANTs~\cite{Avants2008}.

\subsection{Transformer based registration}

Transformers are state of the art networks based on attention mechanism that were first presented for tasks involving machine translation \cite{vaswani2017attention}. Because they effectively reduces the inductive biases of convolutions and can capture  long-range dependencies, this architecture has recently been extensively studied in computer vision applications~\cite{Dosovitskiy2020, Liu2021}. Dual transformer network (DTN)~\cite{Zhang2021} for diffeomorphic image registration was based on a 5-layer U-Net, but this dual configuration consumes a large amount of GPU RAM and significantly raises the computational cost.  Many subsequent  registration methods~\cite{Jia2022,tony, zhao} were motivated by the  success of the  U-Net as the foundation of their registration networks. Some of them~\cite{Jia2022, zhao} used multiple U-Nets in a cascading manner to estimate deformations. Although the number of cascaded U-Nets is constrained by the GPU RAM available, training them on large datasets may not be possible with low-end GPUs, despite the fact that this coarse-to-fine strategy typically results in better final performance.

By incorporating the vision transformer (ViT)~\cite{Dosovitskiy2020} block in a convolutional network designed after the V-Net, \cite{Milletari2016}  introduced the ViT-VNet~\cite{Chen2021}. Instead of using image pairings, they feed encoded image features to ViT-V-Net to minimise the computational cost. Their outcomes were comparable to those of VoxelMorph. By using a more sophisticated transformer architecture (Swin-Transformer~\cite{Liu2021}) as its foundation, the authors of ViT-V-Net~\cite{Chen2021} later refined ViT-V-Net and presented TransMorph~\cite{Chen2022}. They carried out extensive tests and demonstrated the superiority of their TransMorph~\cite{Chen2022} over a number of cutting-edge techniques. In this study, we demonstrate that our suggested technique can perform better on inter-patient and atlas-to-patient brain registration tasks than TransMorph. In this paper we have shown that U-Net architectures with variations are still competitive in the registration of medical images.

\section{Proposed Method}
\label{sec:methods}

\begin{figure*}[thb!]
\centering
\includegraphics[width=\textwidth]{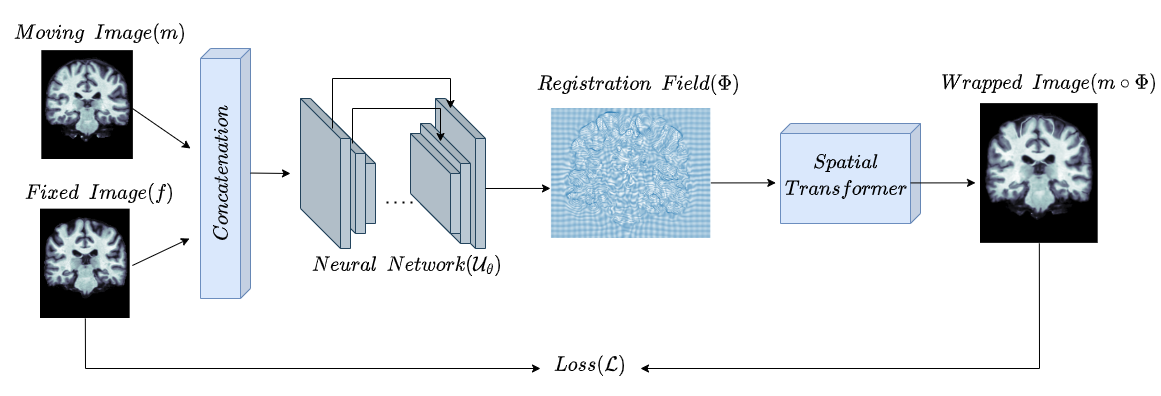}
\caption{Overview of the proposed method. The neural network learns the displacement field $\deform$ to register the 3D moving image to a 3D fixed image. In the process of training, we warp $\mathbf{M}$ with $\deform$ using a spatial transformer function. The loss function compares the similarity between $\moving  \circ \deform$ and encourages smoothness of $\deform$. \label{fig:overview}}
\end{figure*}

In this work, image registration will be treated in the context of unsupervised learning, where we optimize for the parameters in a neural network, whose novel design is the main contribution of the paper. Figure \ref{fig:overview} shows the overview of the proposed method. Essentially the method consists  in minimizing  \ref{eq:unsupervised} described in Section~\ref{sec:apprach} using  the architecture  described in Section \ref{sec:architecture} and the unsupervised loss described in Section \ref{sec:loss}. Detailed descriptions are given in the following.

\subsection{Learning approach}
\label{sec:apprach}

Let $\fixed \colon \Omega \to \R$ and $\moving \colon \Omega \to \R$ be the fixed and moving 3D image volumes that are defined on $\Omega \subseteq \R^3$.  As a pre-processing step, all images are affinely aligned and mapped to the same image domain $\Omega$. In particular, small nonlinear perturbations are the only cause of misalignment between the volumes.

Instead of determining a deformation field $\deform = \id + \disp$ separately for each pair $(\fixed, \moving)$, the learning  approach explicitly constructs $\disp \simeq \net_\theta (\fixed, \moving)$, where $(\net_\theta)_{\theta \in \Theta}$ is parametric function  such as a CNN  and $\theta$ are learnable parameters. In the case that  ground truth  data $(\fixed, \moving, \disp)$ would be available the CNN can defined to minimizes the  supervised loss $\ew[ \ell (\net_\theta(\fixed, \moving), \disp)]$ for some  accuracy  measure $\ell$. Due to the absence of ground truth displacement fields  in most applications we follow here an unsupervised approach. In this case the CNN $\net_\theta$ is defined by
\begin{equation} \label{eq:unsupervised}
\hat \theta = \argmin_\theta  
\ew 
\left[ \fun (\fixed, \moving \circ \deform )  +  \alpha\gun(\deform) \right] \,. 
\end{equation}
Here the expectation value $\ew[\,\cdot\,]$ is taken over pairs $(\fixed, \moving)$ following the  unknown data distribution and $\run_{\fixed, \moving,\alpha}(\deform) = \fun (\fixed, \moving \circ \deform )  +  \alpha\gun(\deform)$ is the variational loss~\eqref{eq:tik}.  The  specific form of the variational loss  using in this work is given in \eqref{eq:loss-concrete} below. Functional~\eqref{eq:unsupervised}  can be minimized by stochastic gradient and variants. Given a new and unseen registration example $(\fixed, \moving)$, computing the registration field is fast and simply consists in one application of the trained network $\net_\theta$.

\begin{figure*}[thb!]
\centering
 \includegraphics[width=\textwidth]{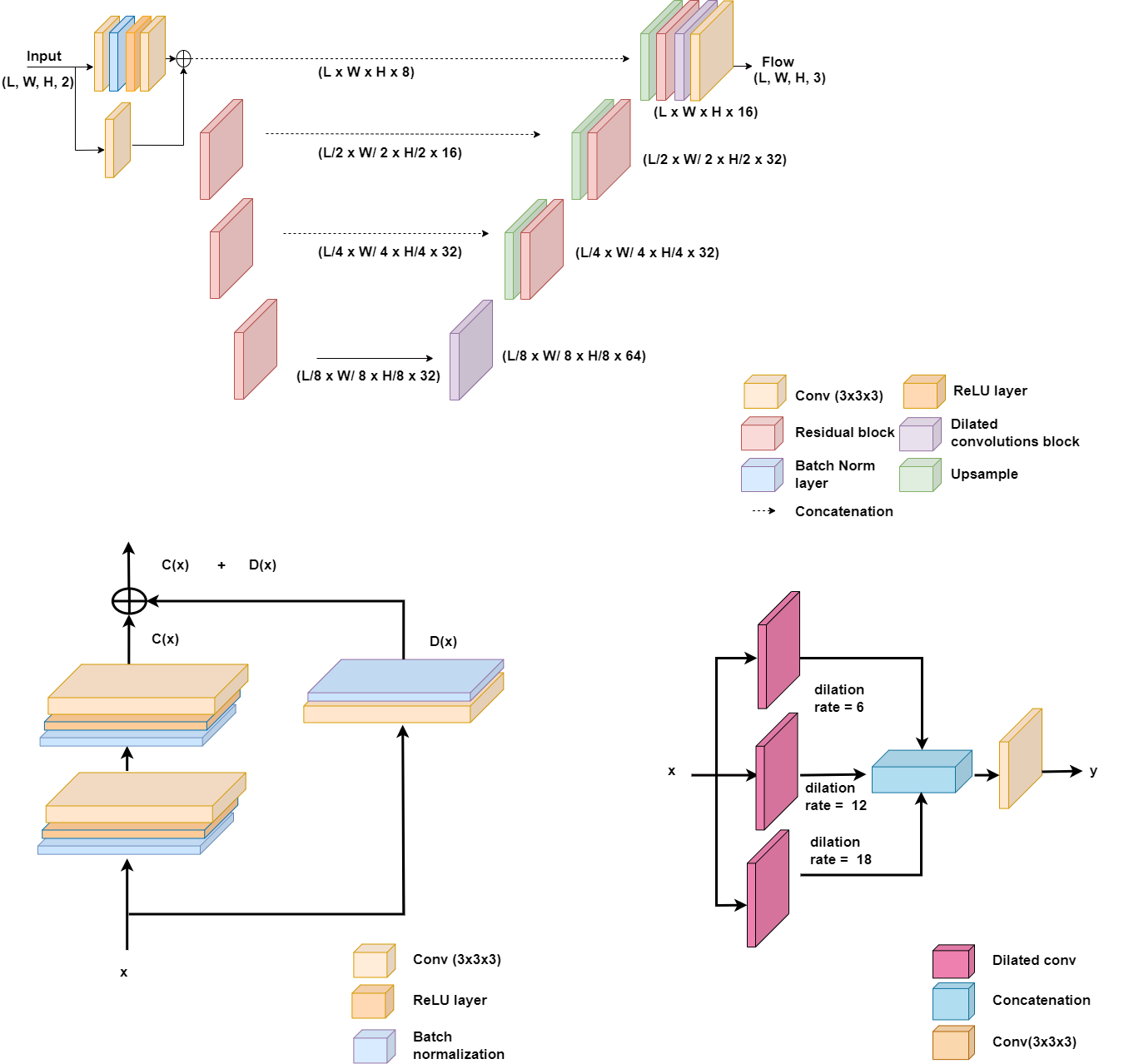}
\caption{Top: proposed Res-Unet with dilated convolution blocks. The network takes stacked moving and fixed images and returns the  displacement field. Bottom left: Used residual block. Bottom right:  Used dilated convolution block.}
\label{fig:blockdiagram1}
\end{figure*}

\subsection{Network Architecture}
\label{sec:architecture}

While theoretically a 5-layer vanilla UNet has a receptive field sufficiently broad to encompass the region potentially affecting the deformation field surrounding a certain voxel, in reality according to RepVGG~\cite{repvgg} and RepLK-ResNet~\cite{ding} the effective receptive field (ERF) of a convolutional network is substantially lower. This is one reason what transformed networks have been proposes for image registration. However, these models are time-consuming to train and evaluate and require a large amount of training data.

To address these challenges, we propose a novel approach in this work: a residual U-Net augmented with embedded parallel dilated-convolutional blocks. This architecture aims to enhance the receptive field while maintaining a compact network size. By incorporating dilated convolutions, our model can effectively capture intrinsic sequence information, broadening the convolution kernel's field without significantly increasing the number of model parameters. Figure \ref{fig:blockdiagram1} illustrates the proposed architecture.

\begin{remark}[Image warping]
In the implementation, images are defined on a Cartesian grid. Therefore, discretization requires $\moving \circ \deform$ warping to the pixel domain. In this work, we use linear interpolation for this purpose, taking
\begin{equation} \label{eq:warped}
    (\moving \circ \deform)(p) 
    = \sum_{q \in Q(\deform(p))} 
    \moving(q) \prod_{d  = 1,2,3}
    \bigl( 1- \abs{\deform_d(p) - q_d} \bigr) \,,
\end{equation}
where $Q(\deform(p))$ consists of all grid points $q$ with $\deform_d(p) - q_d < 1$. As linear interpolation is almost everywhere differentiable, we can use integrate warping  \eqref{eq:warped} in standard backpropagation during training.
\end{remark}

\subsection{Unsupervised Loss}
\label{sec:loss}

The unsupervised loss function used in this work consists of a similarity term $\fun$ that penalizes mismatch between the fixed and the warped moving image, and the regularizer $\gun$ that penalizes local spatial variations in the displacement field. Specifically, the similarity between the fixed image $\fixed$ and the warped image $\warped = \moving \circ \deform$ is taken as the locally normalized cross-correlation (LNCC):
\begin{equation} \label{eq:lncc}
\lncc(\fixed, \warped) =
\\
\sum_p 
\frac{\bigl(\sum_q (\fixed(q)-\fixed_{Z}(p)) \cdot (\warped(q)-\warped_{Z}(p)) \bigr)^2}%
{(\sum_q (\fixed(q)-\fixed_{Z}(p))^2)(\sum_q(\warped(q)-\warped_{Z}(p))^2)} \,,
\end{equation}
where $Z$ is a set of integers defining a neighborhood around $0$. The sum $\sum_q$ is taken over all pixels in the neighborhood of $p$ defined by $p-q \in Z$, and the sum $\sum_p$ is taken over all pixels $p \in \Omega$. Additionally, $\fixed_{Z}(p)$ and $\warped_{Z}(p)$ denote the (local) mean of $\fixed$ and $\warped$ over all $q$ with $p-q \in Z$. In this study, we use $Z=\{-4, -3, \dots, 4\}^3$ to define neighborhoods of $p$ consisting of $9 \times 9 \times 9$ voxels. LNCC is a widely used measure for deformable registration that is particularly robust to intensity differences common across medical images and datasets.

As a regularization term resulting in smooth  deformations $\deform = \id + \disp$, we use the standard quadratic Sobolev norm:
\begin{equation} \label{eq:sobolev}
 \norm{\nabla \disp}_2^2 = \sum_{p \in \Omega} \sum_{i=1,2,3} \norm{\nabla \disp_i(p)}^2 \,,
\end{equation}
where  $\nabla \disp_i(p) \in \R^3$ is the discrete spatial gradient operator defined by finite differences in each dimension. Combining the mismatch and the regularization term results in the unsupervised loss function:
\begin{equation} \label{eq:loss-concrete}
\mathcal{L}_{\fixed, \moving, \alpha}(\deform) = \lncc(\fixed, \moving \circ \deform ) + \alpha \norm{\nabla \disp}_2^2 \,.
\end{equation}
Here, $\deform = \id + \disp$ is the deformation field, $\alpha$ is the regularization parameter, $\lncc$ is defined in \eqref{eq:lncc} and $\norm{\nabla \disp}_2^2$ in  \eqref{eq:sobolev}. The targeted registration function $\disp = \net_\theta (\fixed, \moving)$ is then defined by minimizing \eqref{eq:warped} with the loss \eqref{eq:loss-concrete}.

\section{Experiments and Results}
\label{sec:results}

In this section, we present the results of our studies comparing our method with established ones. We compared our proposed method with six state-of-the-art learning-based methods: VoxelMorph-1, VoxelMorph-2~\cite{Balakrishnan2018, Balakrishnan2019}, ViT-V-Net~\cite{Chen2021}, nnFromer~\cite{Zhou2021}, CoTr~\cite{Xie2021}, and TransMorph~\cite{Chen2022}. Evaluation is done on two tasks, namely atlas-to-patient brain MRI volumes and inter-patient brain MRI volumes. Table~\ref{table:0} shows the number of model parameters and the model size in terms of storage used by the methods.

\begin{table}[thb!]
\centering
\begin{tabular}{ |l|l l|}
 \toprule
Method& Number of parameters   & Size (MB)\\
 \midrule
VoxelMorph-1 & $3 \times 10^{6}$  & 84.9\\
VoxelMorph-2 & $0.3 \times 10^{6} $& 85.3\\
ViT-V-Net   & $110.6 \times 10^{6} $ & 574.1\\
nnFormer    & $45.3  \times 10^{6}$ & 667.6\\
CoTr       & $38.7  \times 10^{6} $ & 685.5\\
TransMorph  & $46.8  \times 10^{6}$ & 815.1\\
Ours        & $0.7   \times 10^{6} $ & 91.6\\
\bottomrule
\end{tabular}
\caption{Number of model parameters and the model size of the networks used in this study.}
\label{table:0}
\end{table}

\subsection{Datasets Description}

\paragraph{Atlas-to-patient brain MRI registration}
For atlas-to-patient registration, we use the IXI dataset containing around 576 MRI scans gathered from three different hospitals. All the scans are from healthy subjects, and we use the pre-processed IXI dataset provided by~\cite{Chen2022}. As we are evaluating the performance on a limited number of samples, we use 30 samples for training, 20 samples for validation, and 115 samples for evaluating our method. The atlas-to-patient brain MRI is obtained from~\cite{Kim2021}. Each volume was reduced to a size of $160 \times 192 \times  224$ pixels, and 29 anatomical structures label maps were used to evaluate registration performance.

\paragraph{Inter-patient brain MRI registration}
For inter-patient registration, we use the OASIS dataset, a public registration challenge dataset obtained from the 2021 Learn2Reg challenge~\cite{Marcus2007, hoopes2021hypermorph}. Specifically, we work with the pre-processed OASIS dataset provided by~\cite{Chen2022} to perform inter-subject brain registration. We use 30 samples for training and evaluate on the provided validation data of 19 samples. Each MRI brain image has a resolution of $160 \times 192 \times  224$ and has had its skull stripped, aligned, and normalized. For evaluation of the registration performance, measures like Dice Score were used with 29 anatomical structures label masks.

\subsection{Results and comparison}

Results on the Atlas-to-patient dataset of our method compared with VoxelMorph-1, VoxelMorph-2, ViT-V-Net, nnFromer, CoTr, and TransMorph are shown in Table \ref{table:1}. To guarantee a fair comparison, we use exactly the same parameters as mentioned in~\cite{Chen2022}. In that table, we compare the average Dice score on validation, Dice score on 115 test samples, and the fraction of voxels with non-positive Jacobian determinant (FNJ).

\begin{table}[thb!]
\centering
\begin{tabular}{ |l|l l l|}
 \toprule
 \multicolumn{4}{|c|}{Atlas-to-patient MRI dataset} \\
 \midrule
Method & Validation  Dice & Test Dice& FNJ\\
 \midrule
VoxelMorph-1 & 0.7050 (0.0280) & 0.7004  (0.0285) & 0.0139 (0.0013) \\
VoxelMorph-2 & 0.7181 (0.0050) & 0.7143 (0.0053) &  0.0149 (0.0001) \\
ViT-V-Net    & 0.6712 (0.0025) & 0.6647 (0.0044) &  0.0180 (0.0004) \\
nnFormer     & 0.7280 (0.0002) & 0.7229 (0.0002) &  0.0119 (0.0001) \\
CoTr         & 0.6134 (0.0280) & 0.6073 (0.0014) &  0.0038 (0.0006) \\
TransMorph   & 0.7250 (0.0001) & 0.7191 (0.0006) &  0.0150 (0.0001) \\
Ours         & 0.7345 (0.0005) & 0.7237 (0.0022) & 0.0158 (0.0004) \\
\bottomrule
\end{tabular}
\caption{Performance evaluation for atlas-to-patient MRI dataset using 30/20/115 samples for training/validation/evaluation.}
\label{table:1}
\end{table}

We see that VoxelMorph-1 has the least number of parameters and achieved a Dice score of 0.70. VoxelMorph-2, with slightly increased parameters, has achieved better performance with a Dice score of 0.71 on the test dataset. Note that some transformer-based methods have the worst performance in the limited data experiment, reflecting that transformers require more data to perform better and be robust. ViT-V-Net has the highest number of parameters but does not perform better than VoxelMorph-1. CoTr comes last in the comparison with only a Dice score of 0.60. The nnFormer and TransMorph have performed very well with Dice scores of 0.72 and 0.719, respectively. Our method, despite using only 0.7 million parameters, has comparable or even slightly better performance than TransMorph and nnFormer with a Dice score of 0.723 on the test dataset.

\begin{table}[thb!]
\centering
\begin{tabular}{ |l|l l l|}
 \toprule
 \multicolumn{4}{|c|}{Inter-patient MRI dataset} \\
 \midrule
 Method& Validation Dice& Test Dice& FNPJ\\
 \midrule
VoxelMorph-1& 0.7533 (0.0009)& 0.7512 (0.0008)& 0.0098 (0.0002)\\
VoxelMorph-2& 0.7623 (0.0026)& 0.7594 (0.0019) & 0.0091 (0.0005)\\
ViT-V-Net&  0.7820 (0.0002)& 0.7782 (0.0001)& 0.0094 (0.0001)\\
nnFormer& 0.7781 (0.0001)& 0.7779 (0.00029 & 0.0008 (0.0002)\\
CoTr&  0.7104 (0.0001)& 0.7102 (0.0002)& 0.0002 (0.0002)\\
TransMorph& 0.7954 (0.0001)& 0.7900 (0.0001) & 0.0066 (0.0003)\\
Ours& 0.8038 (0.0020) & 0.8001 (0.0020) & 0.0008 (0.0001)\\
 \bottomrule
\end{tabular}
\caption{Performance evaluation for the inter-patient MRI dataset using  30/20/115 samples for training/validation/evaluation.}
\label{table:2}
\end{table}

\begin{figure}[thb!]
         \centering
         \includegraphics[width=\columnwidth]{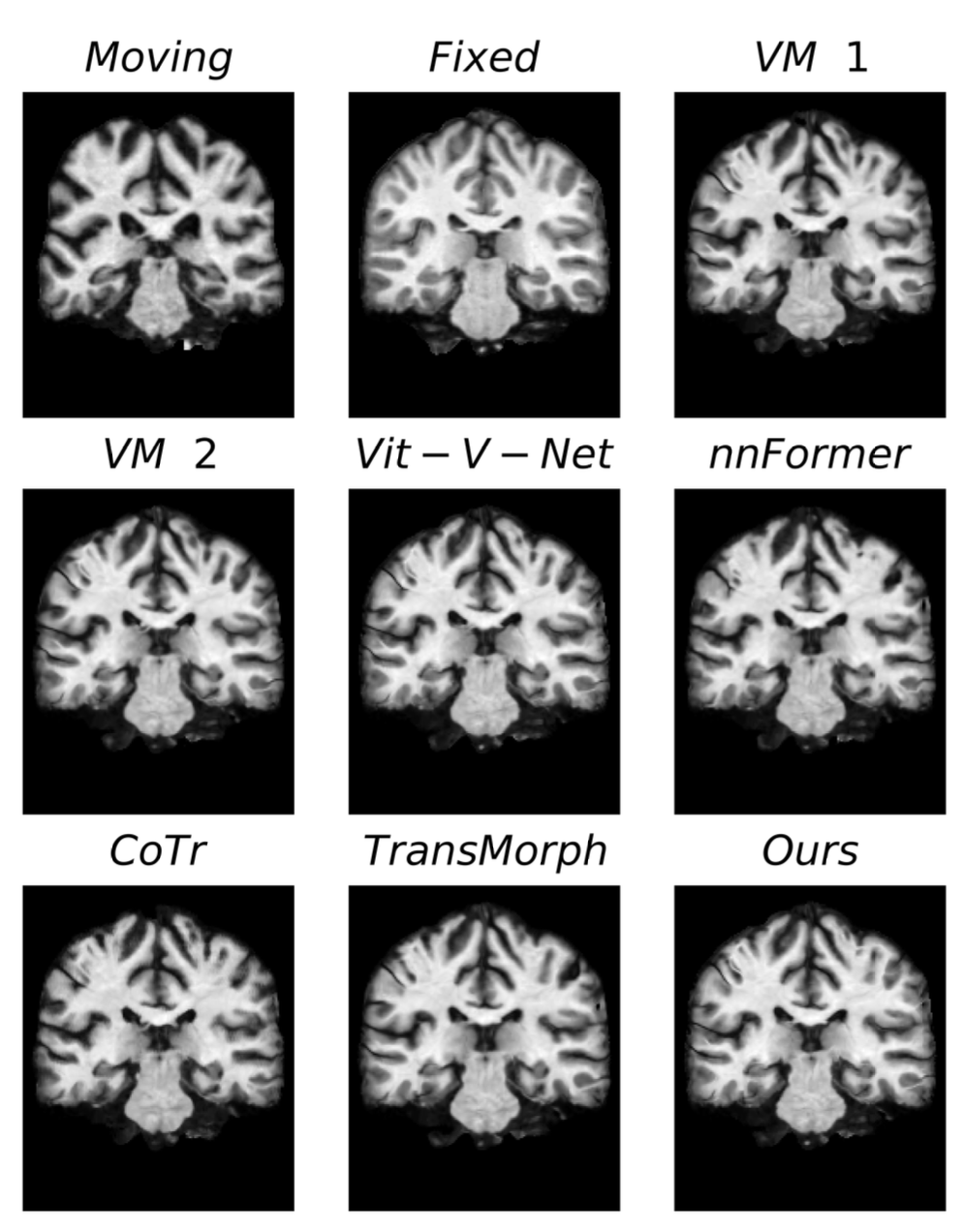}
         \caption{Representative example of registered image  on the atlas-to-patient dataset. \label{fig:ixi}}
\end{figure}

\begin{figure}[thb!]
     \centering
         \includegraphics[width=\columnwidth]{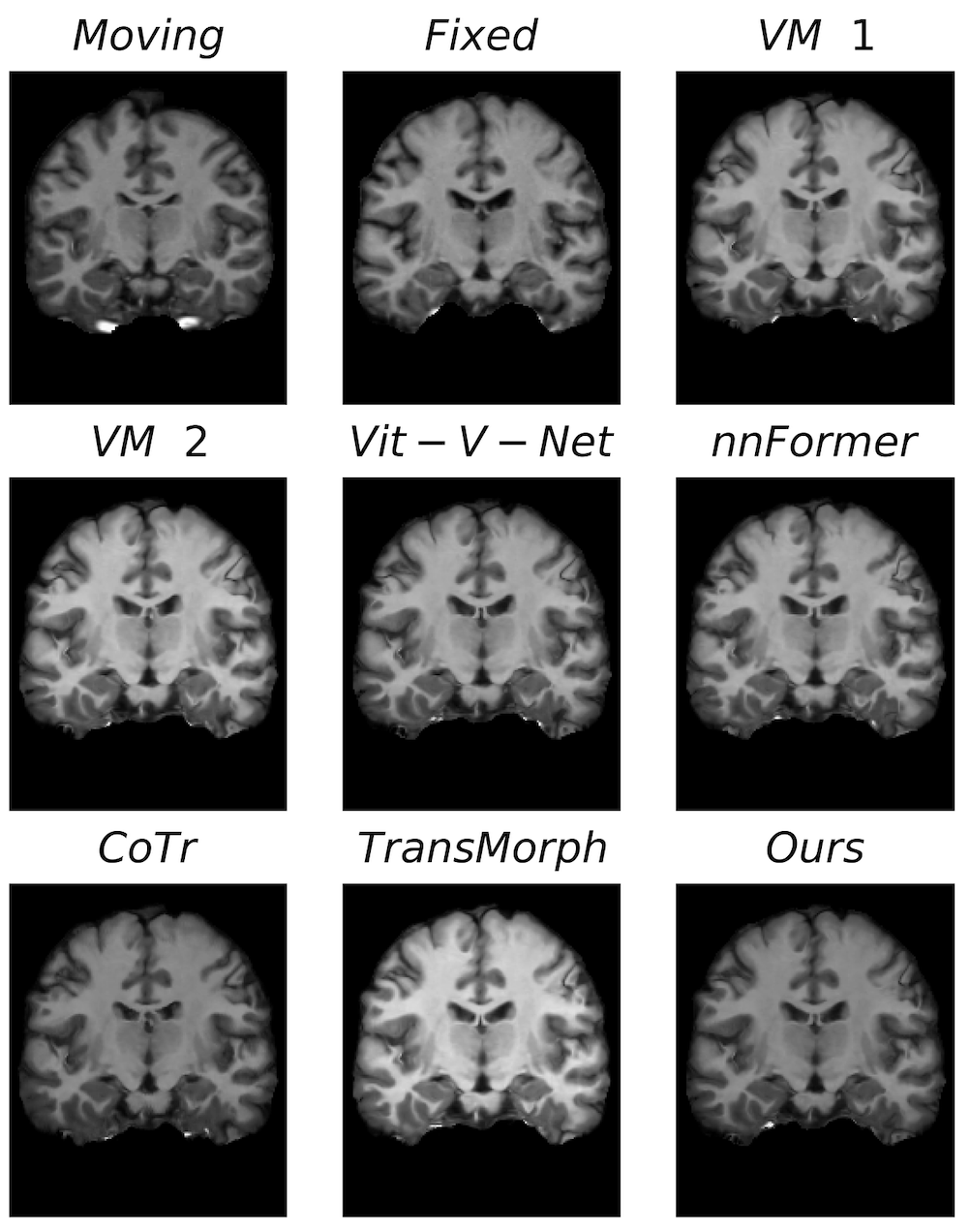}
         \caption{Representative example of registered image in the inter-patient dataset. \label{fig:oasis}}
\end{figure}

\begin{figure}[thb!]
     \centering
         \includegraphics[width=\columnwidth]{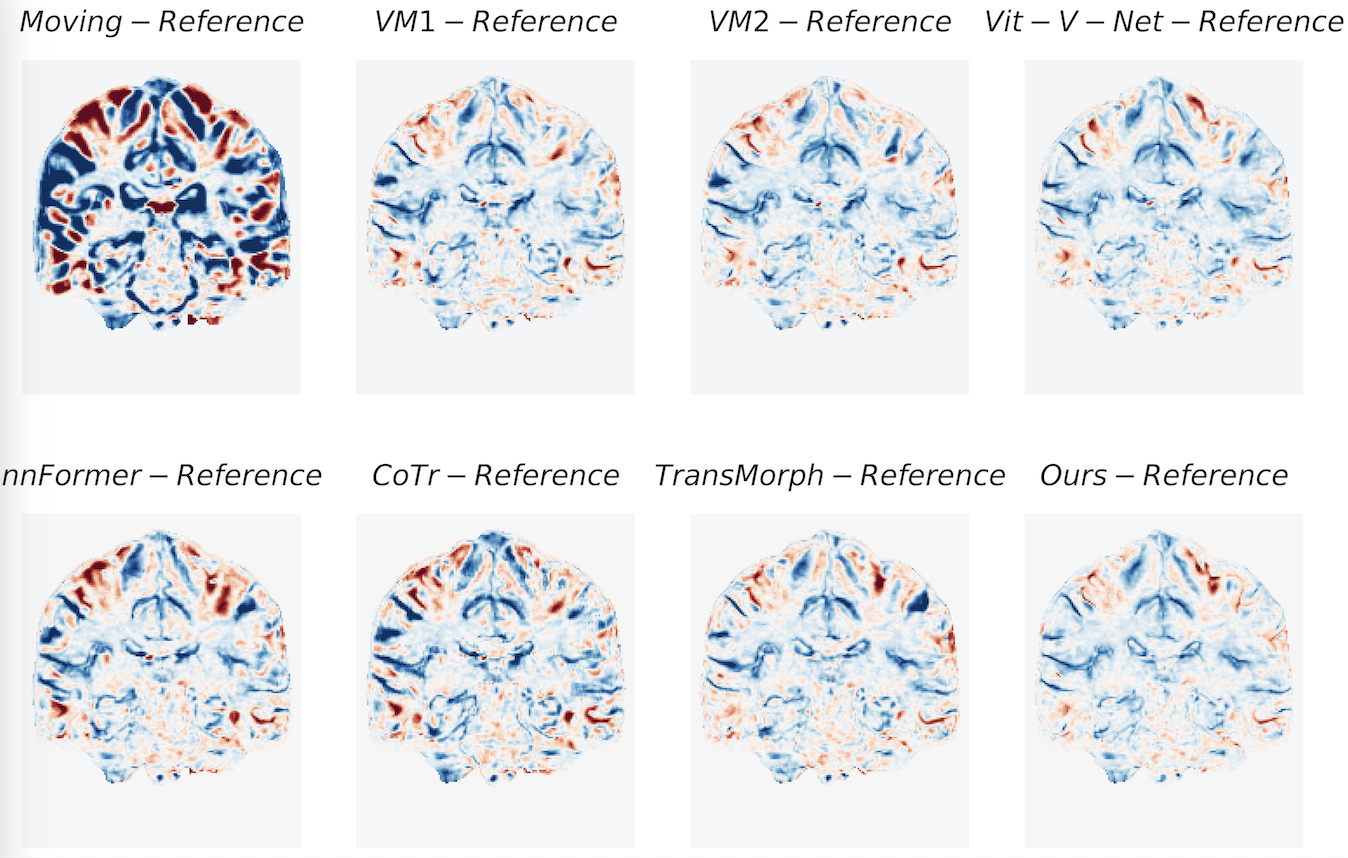}
         \caption{Residual images of a sample volume from atlas-to-patient MRI dataset.}
         \label{fig:ixi-residual}
\end{figure}

\begin{figure}[thb!]
         \centering
         \includegraphics[width=\columnwidth]{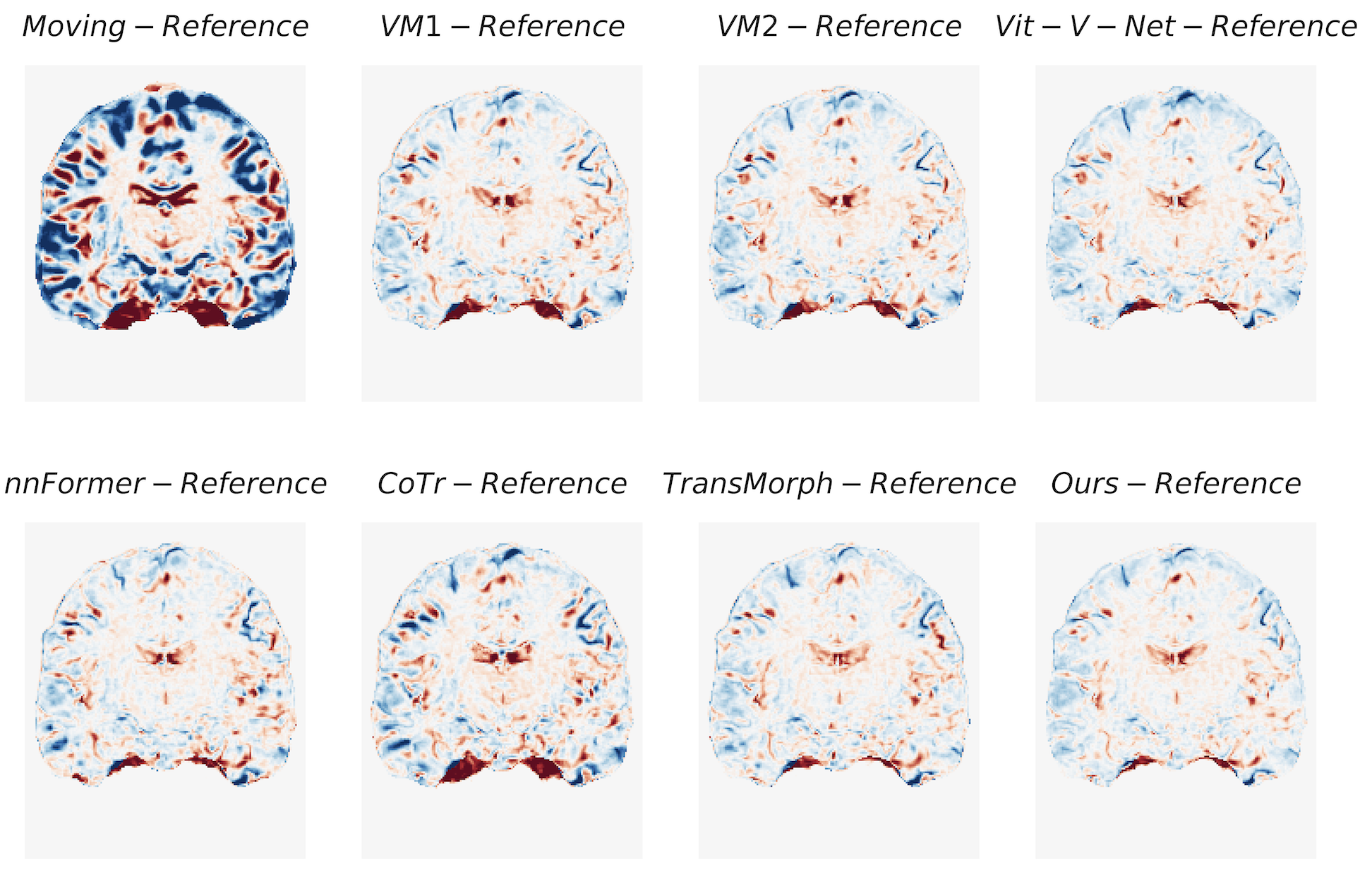}
         \caption{Residual images of a sample volume from Inter-patient MRI dataset.}
         \label{fig:residualt_OASIS}
\end{figure}

\begin{figure}[thb!]
\centering
\includegraphics[width=\columnwidth]{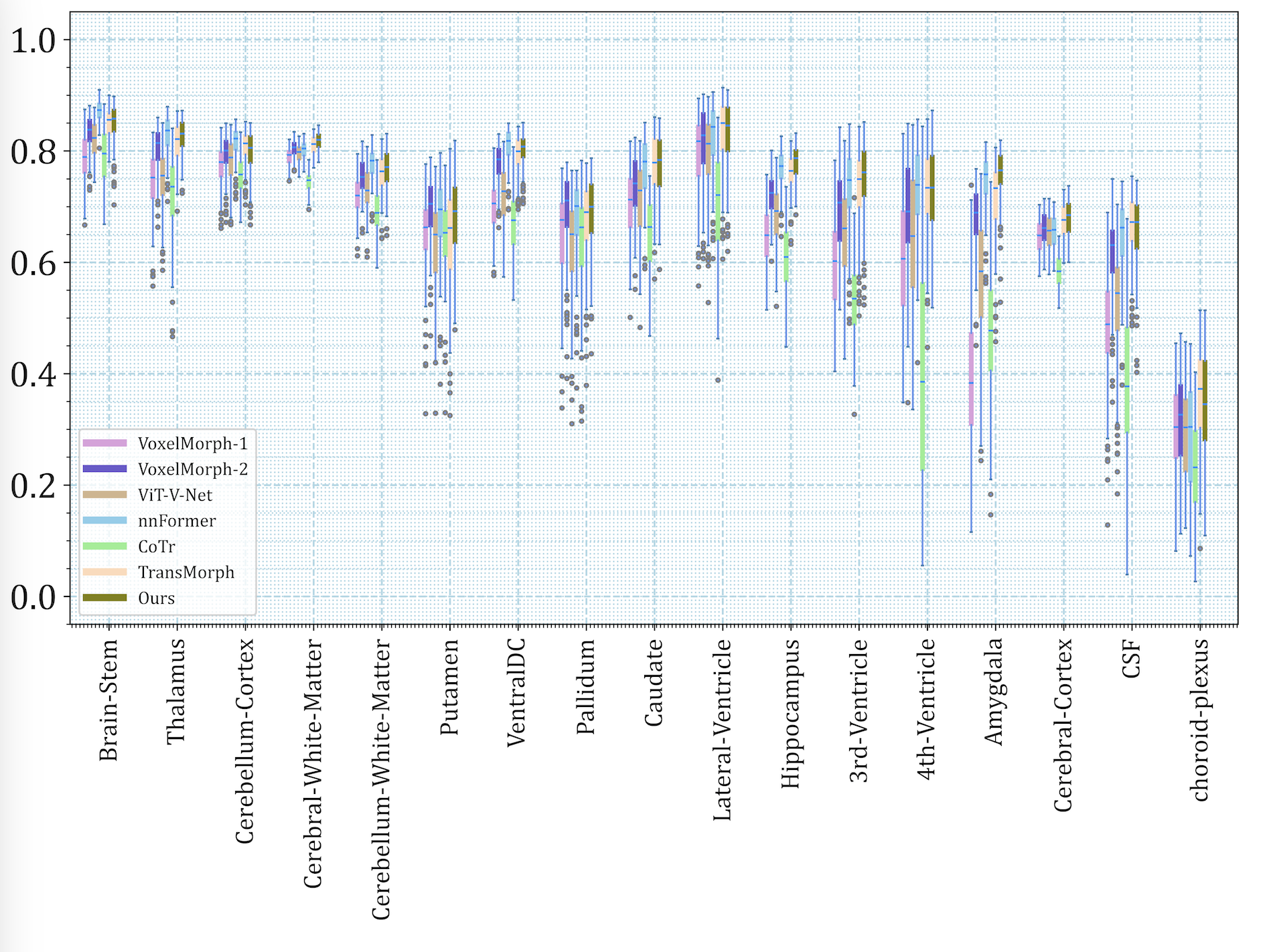}
\caption{Quantitative comparison of various learning-based registration methods. Boxplots showing Dice scores for different organs in atlas-to-patient MRI dataset.}
\label{fig:box_plot_IXI}
\end{figure}

\begin{figure}[thb!]
\centering
\includegraphics[width=\columnwidth]{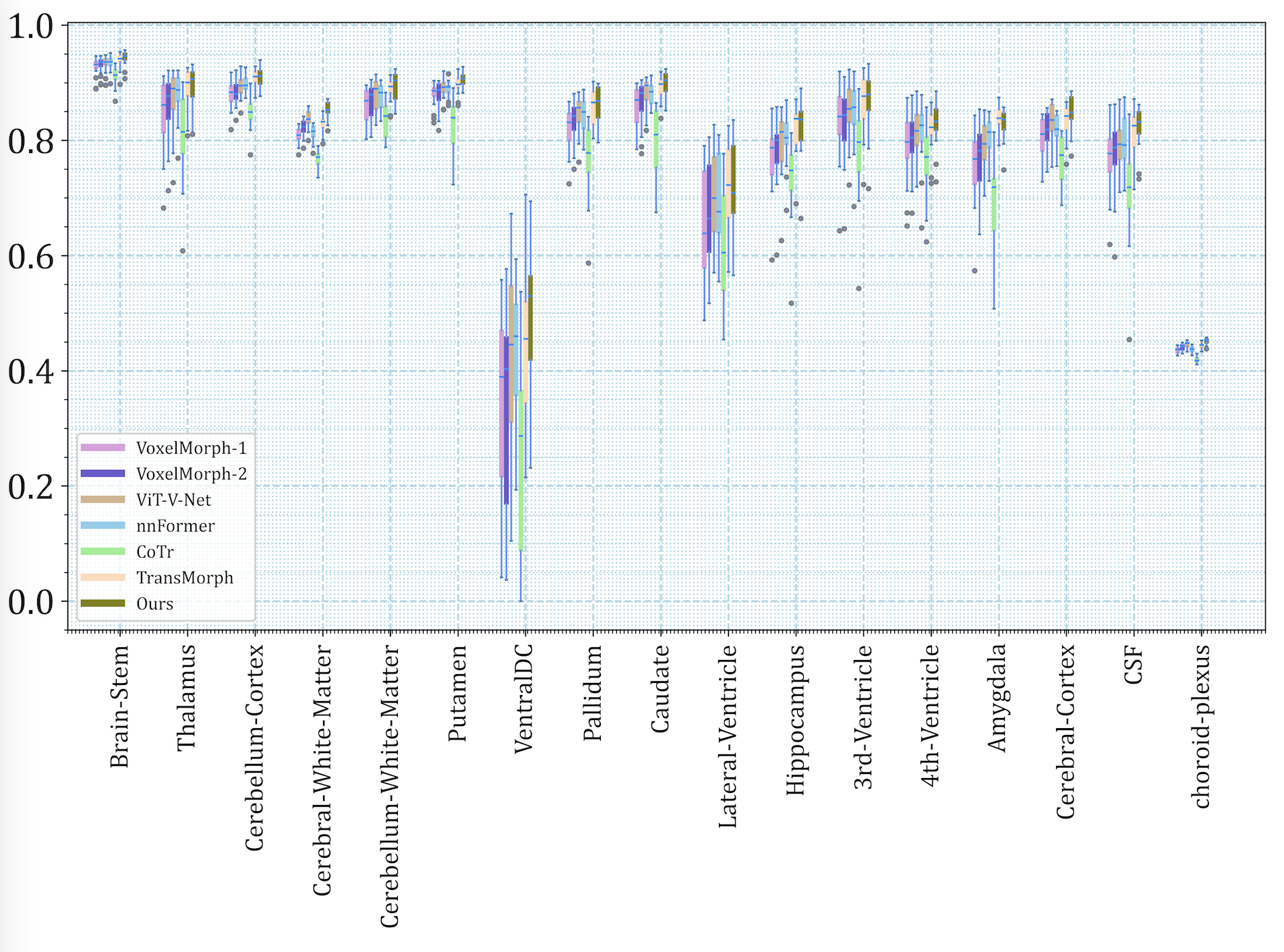}
\caption{Quantitative comparison of various learning-based registration methods. Boxplots showing Dice scores for different organs in inter-patient MRI dataset.}
\label{fig:box_plot_OASIS}
\end{figure}

Similarly, the proposed method has performed better than the other methods on the inter-patient MRI dataset as well, as shown in Table~\ref{table:2}. VoxelMorph-1 has performed the least with a Dice score of 0.75, and transformer-based methods ViT-V-Net, nnFormer, CoTr, and TransMorph have achieved 0.77, 0.77, 0.71, and 0.79, respectively. Our proposed method has achieved a Dice score of 0.80 with a very smaller number of parameters. Figures~\ref{fig:ixi} and \ref{fig:oasis} show representative examples on atlas-patient MRI and inter-patient MRI datasets, respectively. In order to highlight the transformed areas more clearly, Figures~\ref{fig:ixi-residual} and \ref{fig:residualt_OASIS} show the residual images of the MRI volume for atlas-patient and inter-patient datasets, respectively. Finally, Figures~\ref{fig:box_plot_IXI} and \ref{fig:box_plot_OASIS} show boxplots of Dice scores for different organs in atlas-to-patient and inter-patient datasets, respectively.

Transformers are famous for their explicit consideration of long-range spatial relationships. Transformers have been considered as strong candidates for image registration due to their significantly larger receptive field, which allows for a more accurate understanding of the spatial correspondence between moving and fixed images. However, as shown in Table 3, the average lengths of displacements along the $x$-axis, $y$-axis, and $z$-axis are presented. In the atlas-to-patient MRI dataset, our proposed method has achieved larger average displacements in all axes than TransMorph, which are 1.75 voxels, 1.63 voxels, and 1.06 voxels for the $x$-axis, $y$-axis, and $z$-axis, respectively. These displacements are very small compared to the actual image size of 160x192x224. On the inter-patient MRI dataset, our proposed method has slightly fewer average displacements than TransMorph, but the difference is not significant. Therefore, we suggest that it may not be necessary to use transformers, which are bulky, to model long-range dependencies in the deformable image registration task.

\begin{table}[thb!]
\centering
  \begin{tabular}{|l | l l l|}
    \toprule
    Atlas-to-patient & $x$-axis (voxels)& $y$-axis (voxels) & $z$-axis (voxels) \\
    \midrule
    VoxelMorph-1 & 1.37 & 1.09 & 0.48  \\
    VoxelMorph-2 & 1.23 & 1.14 & 0.78  \\
    ViT-V-Net    & 1.13 & 1.11 & 0.75  \\
    nnFormer     & 1.16 & 1.08 & 0.96    \\
    CoTr         & 0.90 & 0.54 & 0.45   \\
    TransMorph   & 1.49 & 1.43 & 1.04   \\
    Ours         & 1.75 & 1.64 & 1.06   \\
    \midrule
    \midrule
    Inter-patient& $x$-axis (voxels) & $y$-axis (voxels) & $z$-axis (voxels) \\
    \midrule
    VoxelMorph-1 & 0.46 & 0.58 & 0.43 \\
    VoxelMorph-2 & 0.50 & 0.56 & 0.45 \\
    ViT-V-Net    & 0.50 & 0.63 & 0.495  \\
    nnFormer     & 0.49 & 0.58 & 0.41  \\
    CoTr         & 0.37 & 0.54 & 0.40  \\
    TransMorph   & 0.64 & 0.81 & 0.66  \\
    Ours         & 0.55 & 0.74 & 0.54  \\
    \bottomrule
  \end{tabular}

  \caption{Average displacements along $x$-axis, $y$-axis and in $z$-axis for all methods on atlas-patient MRI (top) and inter-patient MRI datasets (bottom).}
\label{table:3}
\end{table}

\section{Conclusion}
\label{sec:conclusion}

In this paper, we have shown that the proposed Residual U-Net-based architecture can outperform transformer-based methods. The proposed method has an improved receptive field, few parameters, and significant results on a small training dataset. To improve the receptive field, we suggest Residual U-Net with integrated parallel dilated-convolutional blocks. Atlas-to-patient and inter-patient datasets are used to evaluate the proposed method, and we demonstrate that the performance of the suggested method is similar to or slightly better than transformer-based methods by using only \SI{1.5}{\percent} of its parameters.


\begin{thebibliography}{10}

\bibitem{Ashburner2007}
John Ashburner.
\newblock A fast diffeomorphic image registration algorithm.
\newblock {\em NeuroImage}, 38:95--113, 10 2007.

\bibitem{Ashburner2000}
John Ashburner and Karl~J. Friston.
\newblock Voxel-based morphometry—the methods.
\newblock {\em NeuroImage}, 11:805--821, 6 2000.

\bibitem{Avants2008}
B.~B. Avants, C.~L. Epstein, M.~Grossman, and J.~C. Gee.
\newblock Symmetric diffeomorphic image registration with cross-correlation:
  Evaluating automated labeling of elderly and neurodegenerative brain.
\newblock {\em Medical Image Analysis}, 12:26--41, 2 2008.

\bibitem{Bajcsy1989}
Ruzena Bajcsy and Stane Kovačič.
\newblock Multiresolution elastic matching.
\newblock {\em Computer Vision, Graphics, and Image Processing}, 46:1--21, 4
  1989.

\bibitem{Balakrishnan2018}
Guha Balakrishnan, Amy Zhao, Mert~R. Sabuncu, Adrian~V. Dalca, and John Guttag.
\newblock An unsupervised learning model for deformable medical image
  registration.
\newblock {\em Proceedings of the IEEE Computer Society Conference on Computer
  Vision and Pattern Recognition}, pages 9252--9260, 12 2018.

\bibitem{Balakrishnan2019}
Guha Balakrishnan, Amy Zhao, Mert~R. Sabuncu, John Guttag, and Adrian~V. Dalca.
\newblock Voxelmorph: A learning framework for deformable medical image
  registration.
\newblock {\em IEEE Transactions on Medical Imaging}, 38:1788--1800, 8 2019.

\bibitem{Beg2005}
M~Faisal Beg, Michael~I Miller, Alain Trouv{\'e}, and Laurent Younes.
\newblock Computing large deformation metric mappings via geodesic flows of
  diffeomorphisms.
\newblock {\em International journal of computer vision}, 61:139--157, 2005.

\bibitem{Chen2022}
Junyu Chen, Eric~C. Frey, Yufan He, William~P. Segars, Ye~Li, and Yong Du.
\newblock Transmorph: Transformer for unsupervised medical image registration.
\newblock {\em Medical Image Analysis}, 82:102615, 11 2022.

\bibitem{Chen2021}
Junyu Chen, Yufan He, Eric~C Frey, Ye~Li, and Yong Du.
\newblock Vit-v-net: Vision transformer for unsupervised volumetric medical
  image registration.
\newblock {\em arXiv preprint arXiv:2104.06468}, 2021.

\bibitem{Dalca2019}
Adrian~V. Dalca, Guha Balakrishnan, John Guttag, and Mert~R. Sabuncu.
\newblock Unsupervised learning of probabilistic diffeomorphic registration for
  images and surfaces.
\newblock {\em Medical Image Analysis}, 57:226--236, 10 2019.

\bibitem{ding}
Xiaohan Ding, Xiangyu Zhang, Jungong Han, and Guiguang Ding.
\newblock Scaling up your kernels to 31x31: Revisiting large kernel design in
  cnns.
\newblock In {\em Proceedings of the IEEE/CVF conference on computer vision and
  pattern recognition}, pages 11963--11975, 2022.

\bibitem{repvgg}
Xiaohan Ding, Xiangyu Zhang, Ningning Ma, Jungong Han, Guiguang Ding, and Jian
  Sun.
\newblock Repvgg: Making vgg-style convnets great again.
\newblock In {\em Proceedings of the IEEE/CVF conference on computer vision and
  pattern recognition}, pages 13733--13742, 2021.

\bibitem{Dosovitskiy2020}
Alexey Dosovitskiy, Lucas Beyer, Alexander Kolesnikov, Dirk Weissenborn,
  Xiaohua Zhai, Thomas Unterthiner, Mostafa Dehghani, Matthias Minderer, Georg
  Heigold, Sylvain Gelly, et~al.
\newblock An image is worth 16x16 words: Transformers for image recognition at
  scale.
\newblock {\em arXiv preprint arXiv:2010.11929}, 2020.

\bibitem{droske2004variational}
Marc Droske and Martin Rumpf.
\newblock A variational approach to nonrigid morphological image registration.
\newblock {\em SIAM Journal on Applied Mathematics}, 64(2):668--687, 2004.

\bibitem{He2016}
Kaiming He, Xiangyu Zhang, Shaoqing Ren, and Jian Sun.
\newblock Deep residual learning for image recognition.
\newblock {\em Proceedings of the IEEE Computer Society Conference on Computer
  Vision and Pattern Recognition}, 2016-December:770--778, 12 2016.

\bibitem{Heinrich2013}
Mattias~P Heinrich, Mark Jenkinson, Michael Brady, and Julia~A Schnabel.
\newblock Mrf-based deformable registration and ventilation estimation of lung
  ct.
\newblock {\em IEEE transactions on medical imaging}, 32(7):1239--1248, 2013.

\bibitem{hoopes2021hypermorph}
Andrew Hoopes, Malte Hoffmann, Bruce Fischl, John Guttag, and Adrian~V Dalca.
\newblock Hypermorph: Amortized hyperparameter learning for image registration.
\newblock In {\em Information Processing in Medical Imaging: 27th International
  Conference, IPMI 2021, Virtual Event, June 28--June 30, 2021, Proceedings
  27}, pages 3--17. Springer, 2021.

\bibitem{Isensee2020}
Fabian Isensee, Paul~F Jaeger, Simon~AA Kohl, Jens Petersen, and Klaus~H
  Maier-Hein.
\newblock nnu-net: a self-configuring method for deep learning-based biomedical
  image segmentation.
\newblock {\em Nature methods}, 18(2):203--211, 2021.

\bibitem{Jia2022}
Xi~Jia, Alexander Thorley, Wei Chen, Huaqi Qiu, Linlin Shen, Iain~B Styles,
  Hyung~Jin Chang, Ales Leonardis, Antonio De~Marvao, Declan~P O’Regan,
  et~al.
\newblock Learning a model-driven variational network for deformable image
  registration.
\newblock {\em IEEE Transactions on Medical Imaging}, 41(1):199--212, 2021.

\bibitem{Kim2021}
Boah Kim, Dong~Hwan Kim, Seong~Ho Park, Jieun Kim, June~Goo Lee, and Jong~Chul
  Ye.
\newblock Cyclemorph: Cycle consistent unsupervised deformable image
  registration.
\newblock {\em Medical Image Analysis}, 71:102036, 7 2021.

\bibitem{Lei2020}
Yang Lei, Yabo Fu, Tonghe Wang, Yingzi Liu, Pretesh Patel, Walter~J Curran,
  Tian Liu, and Xiaofeng Yang.
\newblock 4d-ct deformable image registration using multiscale unsupervised
  deep learning.
\newblock {\em Physics in Medicine \& Biology}, 65(8):085003, 2020.

\bibitem{Liu2021}
Ze~Liu, Yutong Lin, Yue Cao, Han Hu, Yixuan Wei, Zheng Zhang, Stephen Lin, and
  Baining Guo.
\newblock Swin transformer: Hierarchical vision transformer using shifted
  windows.
\newblock In {\em Proceedings of the IEEE/CVF international conference on
  computer vision}, pages 10012--10022, 2021.

\bibitem{Long2015}
Jonathan Long, Evan Shelhamer, and Trevor Darrell.
\newblock Fully convolutional networks for semantic segmentation.
\newblock {\em Proceedings of the IEEE Computer Society Conference on Computer
  Vision and Pattern Recognition}, 07-12-June-2015:431--440, 10 2015.

\bibitem{Marcus2007}
Daniel~S. Marcus, Tracy~H. Wang, Jamie Parker, John~G. Csernansky, John~C.
  Morris, and Randy~L. Buckner.
\newblock Open access series of imaging studies (oasis): cross-sectional mri
  data in young, middle aged, nondemented, and demented older adults.
\newblock {\em Journal of cognitive neuroscience}, 19:1498--1507, 9 2007.

\bibitem{Milletari2016}
Fausto Milletari, Nassir Navab, and Seyed-Ahmad Ahmadi.
\newblock V-net: Fully convolutional neural networks for volumetric medical
  image segmentation.
\newblock In {\em 2016 fourth international conference on 3D vision (3DV)},
  pages 565--571. Ieee, 2016.

\bibitem{Modat2010}
Marc Modat, Gerard~R. Ridgway, Zeike~A. Taylor, Manja Lehmann, Josephine
  Barnes, David~J. Hawkes, Nick~C. Fox, and Sébastien Ourselin.
\newblock Fast free-form deformation using graphics processing units.
\newblock {\em Computer Methods and Programs in Biomedicine}, 98:278--284, 6
  2010.

\bibitem{tony}
Tony~CW Mok and Albert Chung.
\newblock Fast symmetric diffeomorphic image registration with convolutional
  neural networks.
\newblock In {\em Proceedings of the IEEE/CVF conference on computer vision and
  pattern recognition}, pages 4644--4653, 2020.

\bibitem{Onofrey2014}
John~A Onofrey, Lawrence~H Staib, and Xenophon Papademetris.
\newblock Semi-supervised learning of nonrigid deformations for image
  registration.
\newblock In {\em Medical Computer Vision. Large Data in Medical Imaging: Third
  International MICCAI Workshop, MCV 2013, Nagoya, Japan, September 26, 2013,
  Revised Selected Papers 3}, pages 13--23. Springer, 2014.

\bibitem{Qiu2022}
Huaqi Qiu, Chen Qin, Andreas Schuh, Kerstin Hammernik, and Daniel Rueckert.
\newblock Learning diffeomorphic and modality-invariant registration using
  b-splines.
\newblock In {\em Medical Imaging with Deep Learning}, 2021.

\bibitem{Redmon2016}
Joseph Redmon, Santosh Divvala, Ross Girshick, and Ali Farhadi.
\newblock You only look once: Unified, real-time object detection.
\newblock {\em Proceedings of the IEEE Computer Society Conference on Computer
  Vision and Pattern Recognition}, 2016-December:779--788, 12 2016.

\bibitem{Ronneberger2015}
Olaf Ronneberger, Philipp Fischer, and Thomas Brox.
\newblock U-net: Convolutional networks for biomedical image segmentation.
\newblock In {\em Medical image computing and computer-assisted
  intervention--MICCAI 2015: 18th international conference, Munich, Germany,
  October 5-9, 2015, proceedings, part III 18}, pages 234--241. Springer, 2015.

\bibitem{Rueckert1999}
Daniel Rueckert, Luke~I Sonoda, Carmel Hayes, Derek~LG Hill, Martin~O Leach,
  and David~J Hawkes.
\newblock Nonrigid registration using free-form deformations: application to
  breast mr images.
\newblock {\em IEEE transactions on medical imaging}, 18(8):712--721, 1999.

\bibitem{scherzer2006mathematical}
Otmar Scherzer.
\newblock {\em Mathematical models for registration and applications to medical
  imaging}, volume~10.
\newblock Springer Science \& Business Media, 2006.

\bibitem{scherzer2009variational}
Otmar Scherzer, Markus Grasmair, Harald Grossauer, Markus Haltmeier, and Frank
  Lenzen.
\newblock {\em Variational methods in imaging}, volume 167.
\newblock Springer, 2009.

\bibitem{sotiras2013deformable}
Aristeidis Sotiras, Christos Davatzikos, and Nikos Paragios.
\newblock Deformable medical image registration: A survey.
\newblock {\em IEEE transactions on medical imaging}, 32(7):1153--1190, 2013.

\bibitem{vaswani2017attention}
Ashish Vaswani, Noam Shazeer, Niki Parmar, Jakob Uszkoreit, Llion Jones,
  Aidan~N. Gomez, Lukasz Kaiser, and Illia Polosukhin.
\newblock Attention is all you need, 2017.

\bibitem{Vercauteren2009}
Tom Vercauteren, Xavier Pennec, Aymeric Perchant, and Nicholas Ayache.
\newblock Diffeomorphic demons: Efficient non-parametric image registration.
\newblock {\em NeuroImage}, 45:S61--S72, 3 2009.

\bibitem{Xie2021}
Yutong Xie, Jianpeng Zhang, Chunhua Shen, and Yong Xia.
\newblock Cotr: Efficiently bridging cnn and transformer for 3d medical image
  segmentation.
\newblock In {\em Medical Image Computing and Computer Assisted
  Intervention--MICCAI 2021: 24th International Conference, Strasbourg, France,
  September 27--October 1, 2021, Proceedings, Part III 24}, pages 171--180.
  Springer, 2021.

\bibitem{Yang2017}
Xiao Yang, Roland Kwitt, Martin Styner, and Marc Niethammer.
\newblock Quicksilver: Fast predictive image registration – a deep learning
  approach.
\newblock {\em NeuroImage}, 158:378--396, 9 2017.

\bibitem{Zhang2021}
Yungeng Zhang, Yuru Pei, and Hongbin Zha.
\newblock Learning dual transformer network for diffeomorphic registration.
\newblock In {\em Proceedings, MICCAI 2021}, pages 129--138. Springer, 2021.

\bibitem{zhao}
Shengyu Zhao, Yue Dong, Eric~I Chang, Yan Xu, et~al.
\newblock Recursive cascaded networks for unsupervised medical image
  registration.
\newblock In {\em Proceedings of the IEEE/CVF international conference on
  computer vision}, pages 10600--10610, 2019.

\bibitem{Zhou2021}
Hong-Yu Zhou, Jiansen Guo, Yinghao Zhang, Lequan Yu, Liansheng Wang, and Yizhou
  Yu.
\newblock nnformer: Interleaved transformer for volumetric segmentation.
\newblock {\em arXiv:2109.03201}, 2021.

\bibitem{Zhu2018}
Bo~Zhu, Jeremiah~Z Liu, Stephen~F Cauley, Bruce~R Rosen, and Matthew~S Rosen.
\newblock Image reconstruction by domain-transform manifold learning.
\newblock {\em Nature}, 555(7697):487--492, 2018.

\end{thebibliography}

\end{document}